# Data Curves Clustering Using Common Patterns Detection

Konstantinos F. Xylogiannopoulos

*Department of Computer Science*

*University of Calgary*

*Calgary, AB, Canada*

kostasfx@yahoo.gr; kxylogia@ucalgary.ca

*Abstract* — For the past decades we have experienced an enormous expansion of the accumulated data that humanity produces. Daily a numerous number of smart devices, usually interconnected over internet, produce vast, real-values datasets. Time series representing datasets from completely irrelevant domains such as finance, weather, medical applications, traffic control etc. become more and more crucial in human day life. Analyzing and clustering these time series, or in general any kind of curves, could be critical for several human activities. In the current paper, the new Curves Clustering Using Common Patterns (3CP) methodology is introduced, which applies a repeated pattern detection algorithm in order to cluster sequences according to their shape and the similarities of common patterns between time series, data curves and eventually any kind of discrete sequences. For this purpose, the Longest Expected Repeated Pattern Reduced Suffix Array (LERP-RSA) data structure has been used in combination with the All Repeated Patterns Detection (ARPaD) algorithm in order to perform highly accurate and efficient detection of similarities among data curves that can be used for clustering purposes and which also provides additional flexibility and features.

## Introduction

Data science is a well-established new science overlapping with many traditional sciences. For decades, statistics is heavily used for standard data science tasks such as detection of patterns or trends in disciplines like financial analysis, medicine, commerce etc. However, with the exploding development of internet, huge amounts of data are produced constantly and more powerful tools have to be developed in computer science and mathematics to help in data analytics, transforming data science to a more concrete discipline. Additionally, a vast number of smart devices have been introduced to our everyday life that have specific characteristics, such as time correlation, leading in the creation of massive time series that need to be analyzed not only correctly and meaningfully but also very fast.

One of the most important tasks that data science deals with is the categorization of our data in order to extract useful information. For example, an emergency room at a hospital needs to categorize its patients based on symptoms before assign a doctor to each case while web retailers group their customers based on their interests and webpage visits. In data science, in order to produce such grouping, two significantly different techniques are used, namly classification and clustering. The core difference between these two techniques is any prior knowledge of information for the dataset that needs to be analyzed. More specifically, classification is used when the classes that we want to group our data are well known and defined in advance. For example, a financial institution may want to classify its customers in order to determine their credit risk. In order to achieve this, financial institutions use specific criteria to determine if a potential loaner has low or high credit risk such as income, age, profession etc. When a new customer asks for a loan the credit risk is calculated based on the criteria that the bank has set and classified as low or high credit risk. However, when the classes are not known in advance then clustering is used. This process is usually significantly harder because of the absence of any previous knowledge on how to categorize our data. A classic example is object recognition. If we have photos from different objects, for example, fruits, cars, humans etc. we care to cluster these objects in such a way that objects of the same cluster are more similar among each other compared to objects from another cluster.

What makes classification and clustering extremely difficult tasks is the dimensionality of the data points or elements description, i.e., the number of attributes that describe the element. The more attributes, the more difficult is to successfully group the elements, a problem described in data science as the curse of dimensionality. Because of this, it is fundamental to apply a dimensionality reduction in order to simplify our data with the least possible information loss. However, there are cases that the structure of our data either does not allow us to do it or they are so complicated that any dimensionality reduction is useless. An example of such case is curves. Curves can be produced by any kind of data related to time such as financial time series, weather data, traffic control etc.

The analysis of time series or in general curves, can be very important in order to detect trends, structures, patterns etc. For example, in weather data analysis it is crucial to be able to analyze huge historical data on variables such as ground and air temperature, humidity, air speed etc. Although such kind of analysis is very helpful for everyday life, yet, it can be extremely important in activities such as aviation when a potential flight redirection could avoid a disastrous accident because of a thunderstorm. It is obvious that our ability to analyze such time series and produce meaningful results is an important task that needs to be executed quickly and accurately. Usually, such kind of time series datasets, formulate the 3 Vs of Big Data Analytics, i.e., volume, velocity, variety. Three Vs define big data based on the size they have, how fast they are produced and the diversity of information they accumulate. Furthermore, as we mentioned before, dimensionality and its problems are very important in such analyses. For example, when a new patient arrives at a hospital, the severity of his/her condition can be defined based on several attributes and the patient should classified based on those which are critical for his/her current state. Although several attributes can be used for such classification, a patient represents a single, multidimensional, data point. However, the classification of time series is significantly more difficult. A time series and its equivalent curve, although it has usually only 2 dimensions it cannot be classified as easily as it is expected. A time series is a collection of 2-dimensional data points strictly ordered together by the time factor. Therefore, its internal structure cannot be bypassed when we need to analyze it.

A classic example is the classification of financial time series. In stock markets we have thousands of stocks and indices that we need to analyze in order to increase our profit or reduce our risk. There are several tools that try to perform forecasting and recommendation based on the structure of a stock, i.e., its diagram shape. More specifically, Technical Analysis relies heavily on such an analysis and one of the most common tools for decades is regression analysis which can shows the trend of a stock represented by its curve. However, such analyses can be very miss-informative. For example, two stocks at a specific point in time can have exactly the same regression analysis characteristics, yet, a completely different shape, which could probably mean that one could continue in up-trend and the other in down-trend. In order to group together different stocks that have the same performance in general the most appropriate way would be to map their curves.

In this paper a new approach is presented that allows us to perform curve clustering using the shape of the curves and common pattern that appear at same locations between curves. The new methodology is an extension of Big Data Curves Clustering (BD2C) with significant improvements and differences. More specifically a text mining and pattern detection approach will be presented based on the Longest Expected Repeated Pattern Reduced Suffix Array (LREP-RSA) [21-22] data structure and the All Repeated Patterns Detection (ARPaD) [20, 22] algorithm. LERP-RSA has some unique attributes, described in the next section, and in combination with the ARPaD algorithm, a multivariate pattern detection methodology will be presented which will allows us to produce a sufficient, accurate, as much as possible, and fast solution for the curve clustering problem.

The rest of the paper is organized as follows: Session II presents related work, relevant data structure and algorithm which will be used in the methodology introduced. Section III presents the necessary steps of the methodology in order to solve the curve clustering problem. Section IV presents examples of methodology usage and the experimental analysis. Finally, Section V presents conclusions and future applications of the proposed methodology.

# Related Work

## Time Series Clustering

Time series clustering is always an important problem with significant attention. In their very wide-ranging and detailed review, Aghabozorgi, Shirkhorsidi and Wah, define the problem of time series clustering as the unsupervised process of partitioning a dataset of $n$ time series $D$ into $C$ homogeneous groups based on a certain likeness degree without any prior knowledge regarding groups definitions [1, 12]. This definition can be expanded to any curve representing any type of sequence. There are two types of sequences a discrete sequence, usually named temporal sequence, and a real values sequence, usually named time series [1]. The problem of clustering time series or in general real values sequences is considerably more difficult than discrete sequences. In the current paper it will be presented how this problem can be reduced to discrete sequence clustering and solved using already developed, state of the art, techniques [22].

The problem of clustering continuous sequences is usually very challenging because, from the one hand, time series size is big (and systems of time series further increases the problem size) and, from the other hand, usually they are stored on disks, which increases exponentially the clustering difficulty [1]. Furthermore, the high dimensionality of the time series plays an important role in the hardness of their clustering [1, 6, 7]. In order to address dimensionality problem, dimensionality reduction methods are used to accelerate the analysis performance [7]. Another reason that can increase the level of difficulty is the choice of a similarity measure [1]. Additionally, what can make almost impossible in many cases curve clustering is the presence of noise in the dataset with the existence of outliers and shifts [9].

In literature, three categories of time series clustering are described. The first category is the whole time series clustering based on the similarities that exist between time series [1]. The second category is the data point clustering where the clustering is related to the temporal proximity and the similarity of the data points, while the last category is subsequence clustering on parts of time series using a sliding window technique [1]. It needs to be mentioned though that according to Keogh and Lin [8] such clustering is practically meaningless.

As mentioned already, dimensionality reduction is a crucial factor for time series analysis and clustering. An important parameter of time series clustering related to dimensionality reduction is the representation of the time series. This kind of representation is divided into four different forms. The first is the Model Based which uses stochastic methods such as Markov models and hidden Markov model [1, 11], statistical models, Auto Regressive Moving Average [1, 5] etc. The second form is the Data Adaptive where the main effort focuses on the minimization of the reconstruction error by applying random length slices [1, 17], such as Singular Value Decomposition [4] etc. A dissimilar approach to the data adaptive is the Non-Data Adaptive methods such as Discrete Wavelet Transform [1]. The last form is based on Data Dictated methods, which in contrast to the other three, it uses a compression ratio which is calculated automatically based on the original data [13].

Several algorithms related to time series clustering exist and are classified in six categories. The first category is the Hierarchical clustering algorithms which work in two directions. Either in bottom-up by considering each time series a cluster and progressively merging clusters to a larger one or top-down by starting with one cluster constructed from all time series and gradually break it down to smaller clusters [1]. Examples of hierarchical algorithms are the Dynamic Time Warping, Discrete Time Warping or its improvements such as the Uniform Time Warping [14, 15, 19] and the Longest Common Subsequence [16, 2]. Another category is the Partitioning Clustering algorithms such as the k-Means [10] which tries to minimize the distance of the elements of a cluster from the centroid. The third category is the Model-based clustering where the main goal is the reconstruction of the original model from a subset of data. The process is based on the assumption of a model for each cluster and then the attempt to best fit the data to that model [1]. The fourth category is the Density based clustering algorithms with the famous DBSCAN [3]. In this case clusters are created based on their high density and separated from other clusters due to the low-density space between them. The last two categories are the Grid-based and the Multi-step. The first divides the space to a finite set of cells forming a grid while the second use several different steps from other techniques as hybrid systems and trying to optimize the outcome [1].

## Repeated/Common Patterns Detection

As mentioned earlier the main target of the methodology is to perform curve clustering using repeated patterns that are common between curves, i.e., exist at the same positions. In order to achieve this, first we need to transform time series or any kind of data representing curves to discrete sequences. Discretization of continuous values is a big topic in data mining. Several methods have been developed in order to allow such transformation. They can be classified mainly as Primary, when the discretization is achieved without the use of any other method, or Composite when a primary method is initial used for a first discretization and then a second method is used for adjustments [18]. There are several taxonomies such as Supervised vs Unsupervised based on the use of class information, Parametric vs non-Parametric based on user inputs, Hierarchical vs non-Hierarchical by applying an incremental selection of data points, Fuzzy vs non-Fuzzy where a quantitative feature is used initially for discretization and then a membership function is used over the cutting points [18]. For the selection of the cutting points there are several methods based on the size and distribution of the data points. The most common methods are the Equal Width where the space of the observed values between minimum and maximum is cut in same size classes and the Equal Frequency where the space is cut in classes that have exactly the same number of data points [18]. More sophisticated methods exist that can use statistical measures of dependency, such as Chi-Square statistic, fuzzy logic techniques with the application of a membership function, decision tree similar classification rules, such as ID3, etc. [18]

When the first phase of converting curves to discrete sequences is completed then the next phase is executed which is the creation of a special data structure, the Longest Expected Repated Pattern Reduced Suffix Array (LERP-RSA). The LERP-RSA is used in order to allow the pattern detection algorithm All Repeated Patterns Detection (ARPaD) to be executed. Although LERP-RSA is based on the suffix array data structure [20], however, it has significant differences. LERP-RSA uses the actual, lexicographically sorted, suffix strings that are created from a discrete sequence, combined with the suffixes position, while the original suffix array uses only the indexes of the strings and the sequence [20, 22-23]. This may seem that creates a quadratic space complexity data structure, however, it has been proved by the Probabilistic Existence of Longest Expected Repeated Pattern Theorem that the space can be only *O(nlogn)* [22-23]. More precisely, instead of using the whole suffix strings, the theorem allows us to use only the LERP long part of each suffix string and, therefore, significantly reduce the size of the data structure [22-23].

Moreover, LERP-RSA has some exceptional attributes, compared to other data structures used for data mining and pattern detection. First of all, based on the alphabet that has been used to discretize a sequence, it allows the self-classification of the data structure [22-23]. Therefore, hundreds or thousands of classes can be created that are significantly smaller than one-block data structure. This unique characteristic is very important because it provides another core attribute, which is the network and cloud distribution of the data structure with absolute isolation between classes. Since the subclasses created are based on the alphabet this means that patterns cannot overlap between different classes. Therefore, we can spread classes over different network or cloud locations and analyze each one individually. Another attribute is the parallelization which is achieved based on the aforementioned classification [22-23]. Since there are many isolated classes, each one can be sorted independently and although a *O(nlogn)* sorting algorithm has to be used, such as Merge Sort, yet, the sorting is not on the full size of the sequence *n* rather than is on the size of the classes. Moreover, additional algorithms that need to be executed on each class, such as ARPaD, can eb executed in parallel and significantly reduce overall run time. An additional attribute is that the data structure supports self-compression with the use of classification which further reduce its size and accelerates the access process [22-23]. However, another very important and unique attribute for the current methodology is the creation of multivariate LERP-RSA data structure that can support more than one sequence [23]. In this case multiple LERP-RSA data structures from many sequences can be combined to create a large LERP-RSA that can be analyzed and provide repeated patterns that (a) exist in the same sequence (same as single LERP-RSA), (b) repeated patterns that exist both in single and different sequences and (c) repeated patterns that although exist only once in sequences, they are repeated (common) patterns between two or more sequences. This feature will be explained in detail in the next section.

For the pattern detection ARPaD algorithm is used. The specific algorithm has been proved to be very powerful, adaptable, fast and with applications in many diverse and completely unrelated fields. Based on literature review ARPaD is probably the only algorithm that can perform all repeated patterns detection, regardless of the frequency of appearance, i.e., without the commonly used support feature of frequent patterns data mining algorithms. ARPaD algorithm has *O(nlogn)* worst case time complexity while in many cases it can be on average *O(n)* [21, 23]. Furthermore, since it executed on the LERP-RSA classes, this means that parallelism and distributed execution can also be applied, which allows increased performance with limited resources. Another important attribute of ARPaD is that it can take many initial parameters to perform targeted pattern detection. For example, it can take different Shorter Pattern Length (SPL) and LERP initial values in order to detect patterns having specific length [21, 23]. In [24] where the BD2C methodology was introduced, a simple version of the algorithm was used to cluster curves. More precisely, in the simple version we set SPL and LERP to have the same value, one. Therefore, we detect single character patterns and based on metanalyses we can measure the similarity of curves and clustered them together. Additionally, qualitive parameters can be defined such as the appearance of a specific alphabet element in a predefined position etc.

# Methodology

The methodology described here for the clustering of data curves is divided in five steps in the same way as BD2C. The first step is the standardization of the data values, which is important for the alignment and scaling of the curves. Then the discretization step is executed in order to transform continue values to discrete based on a predefined alphabet. The third step is the creation of the multivariate LERP-RSA data structure which allows to execute in the fourth step the ARPaD algorithm and detect repeated patterns among different time series. Finally, in the fifth step an analysis of the results produced by the ARPaD is performed in order to identify possible similarities and, therefore, clusters among the curves. The process is described in details in the next paragraphs.

## Dataset Standardization

The first step that is important to perform is standardization of the dataset values in order to align curves and make possible comparison of their shapes. In many cases the curves that we need to cluster or compare, do not have the same units or may not be aligned on an axis (Fig.1). For example, if we want to cluster stock prices to stock indexes then we face two problems, (a) stock indexes do not have units and if we compare stock prices from different stock exchanges the values may refer to different currencies and (b) data values may have significant absolute value difference. For example, Nasdaq-100 Index (NDX) has currently a value close to 7,000 while Apple stock worth approximately 200 USD in New York and 160 EUR in the Frankfurt Stock Exchange. In order to compare and cluster index and stock curves we need first to standardize the dataset in a way that the curves are aligned, yet, the shape is retained.

There are several methods that allows us to standardize values such as the logarithmic function or division by a constant value. However, in our case we care to both align the curves and scale them down in a way that allows us to compare them and, therefore, cluster them in the best possible way. In order to achieve this, the Z-Score transformation should be executed on the dataset values using the formula

$$z = \frac{x - \mu}{\sigma}$$

where $x$ is the actual value, $\mu$ is the mean of the dataset and $\sigma$ is the standard deviation, while z is the output value of the transformation. The benefits of this transformation are two, (a) by subtracting the mean value, the transformed values have mean zero and by dividing by the standard deviation the transformed values have standard deviation one. Therefore, by applying this transformation on two different curves they will have after the transformation the same mean and standard deviation. Both curves will be centered around the same standard value and have the same dispersion since practically z-scoring first moves the curve by a value $\mu$ and then it scales down the curve by a value $\sigma$. Moreover, the shapes of the curves are preserved (Fig.2) and therefore are prepared for further analysis and clustering.

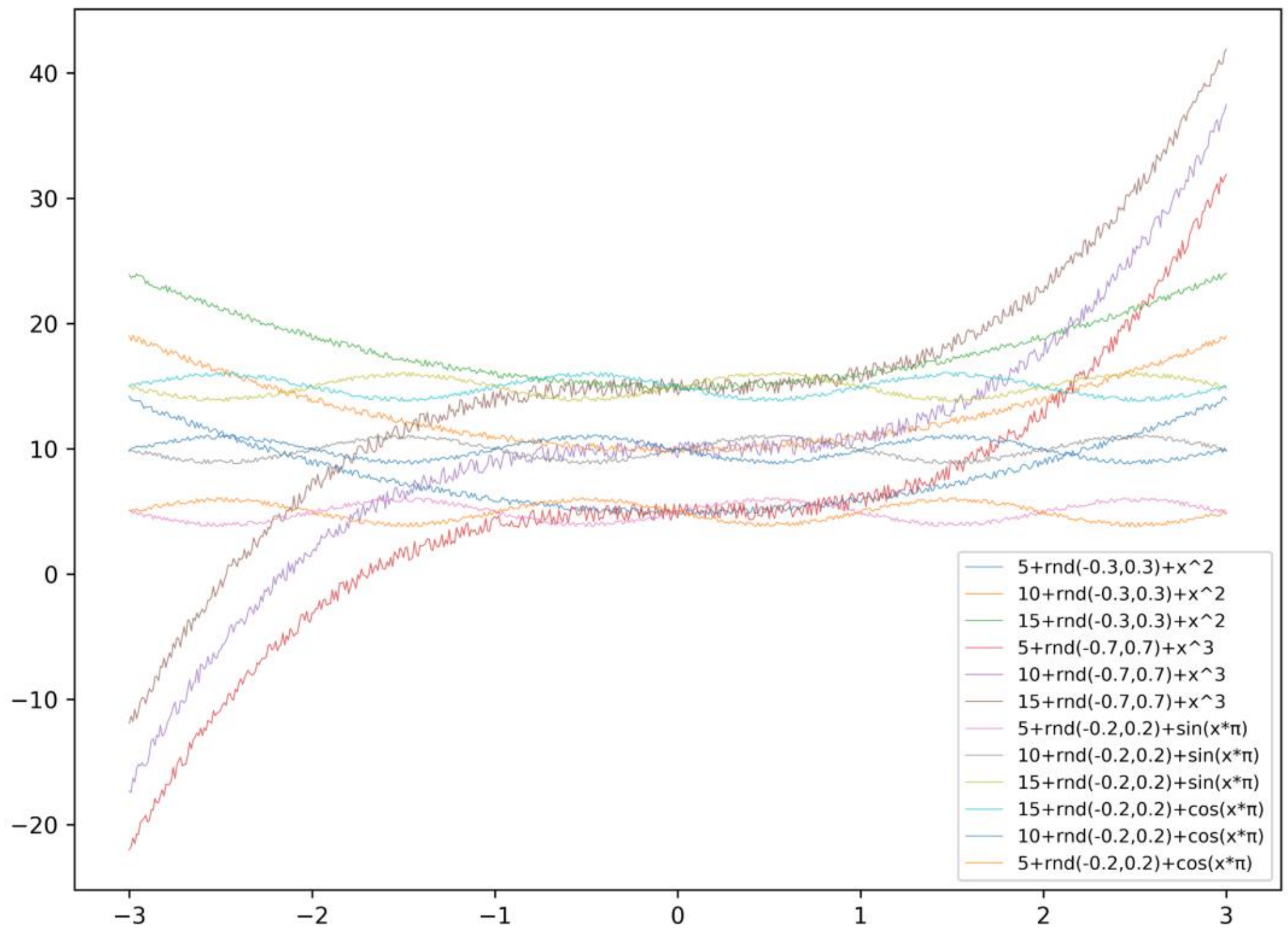

Fig. 1 Original artificial curves with noise

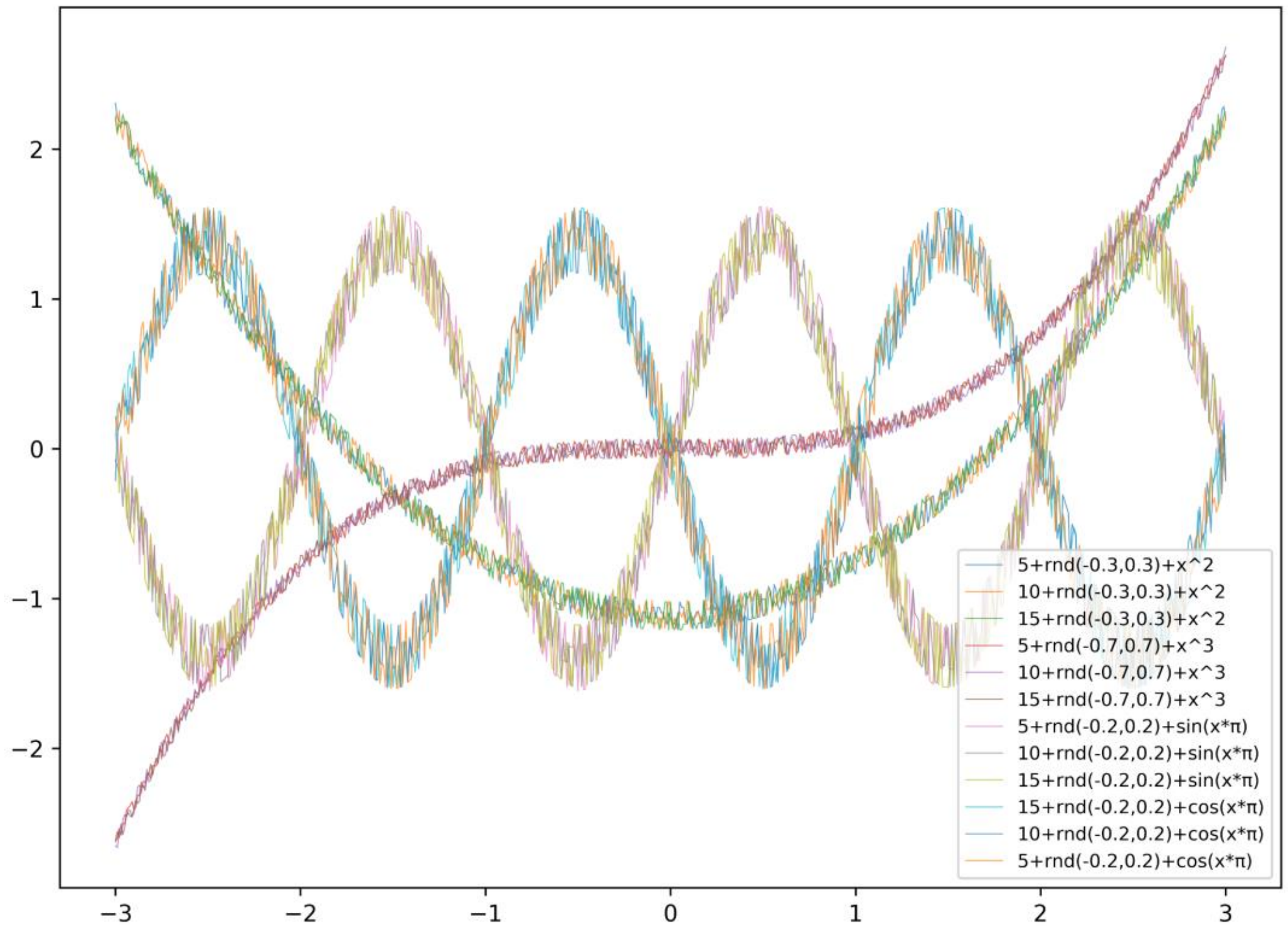

Fig. 2 Z-Scoring transformation on artificial curves with noise

Another issue that sometimes arise during standardization is the possible miss-configuration of the curve length. This problem is very similar to sequence alignment in bioinformatics where possible insertions or deletions may occur in a DNA or RNA sequence. In this case, we may end up with sequences of different lengths, which we need to align in order to perform any pattern detection and clustering based on their shape. This could happen, for example, with meteorological time series where weather sensors may work under extreme conditions and limited hardware resources, which do not allow us to properly record a time series. In this case we may record only a sequence of values without the corresponding time stamp in order to save space, something very common to old sensors and measurements already exist. However, we may face problems that could reduce by a few values the time series, for example, because of an extreme solar activity that could block the device or expand the length of the time series by a few points because of noise duplicates during transmission. Usually, such deformations are random and small in comparison to the actual length of the time series, making them impossible to be detected and fixed. In cases like these we can apply a similar approach to Uniform Time Warping and try to eliminate them. This elimination works as follows: First, we define the expected length of the sequences. Then we identify any sequence which is longer or shorter. For each one of the deformed sequences we calculate the Least Common Multiple (LCM) of its length with the expected length of the sequence. Then we expand the sequence to the length of the LCM by using cubic interpolation between the actual points. When the new, transformed, sequence has been created we can reduce its size to the actual by down sampling the sequence. To achieve this, we select data points starting at first position and with a step equal to the fraction of the LCM over the actual curve length. With this method although we might have a small information loss, however, we maintain the shape of curve, which is the most important benefit of this transformation for the specific methodology.

For example, let us assume that we have three sequences with length 8, 10 and 12 values and that the actual (expected) length is 10. Therefore, we have one sequence that is shorter (8 values) and one that is longer (12 values). Furthermore, it is impossible to determine the positions of the missing values for the first and the positions of the inserted values for the last sequence. In order to properly align the sequences after the z-scoring we need to calculate the LCM for each sequence with the normal recorded one (40 for the first and 60 for the last). Now we expand the sequences, by 32 values the first one and 48 the last, by applying interpolation between the actual recorded values. After that, we have two new sequences with lengths 40 and 60 which we can easily downsize to 10 values by one quarter the first and one sixth the last. The specific method can be applied even if we do not have a standard, actual, length by selecting an arbitrary length such as the mean of lengths between sequences. Then we can transform all sequences to have the same length.

An example with real data we can see in Fig. 3 and Fig. 4 where we have the same time series representing the air temperature measurements in Athens, recorded every six hours, for a total time span of 2,000 data points. In both diagrams the original time series is in orange color. The difference between the two diagrams is that the blue curve represents a 50% shorter time series because data points have been eliminated due to external factors. In the first case we have a uniform removal of data points (even values), while in the second a random removal. On the deformed curves we apply Cubic Spline interpolation and then we downsize the curve to original length of 2,000 data points (green line). As we can observe the shape has been reserved for the new curve in comparison to the original in both case, uniform and random reduction.

It is important to mention that the y-axis values do not represent actual temperatures except for the blue curve in Kelvin units. All other curves have been shifted above in order to avoid overlapping which would make the diagrams confusing. Another very important note is that the 50% reduction (1,000 data points) is huge for real data. Despite that, in the uniform reduction we can observe that the curve has been reconstructed very well, while this is not the case with the random. This happens because the random data points removal might eliminate large, continuous regions of 50 or more data points which are difficult to be precisely reconstructed with interpolation. That is why we can observe that although the cubic spline curve maintains the general shape of the original curve, yet, in specific areas there are some dissimilarities.

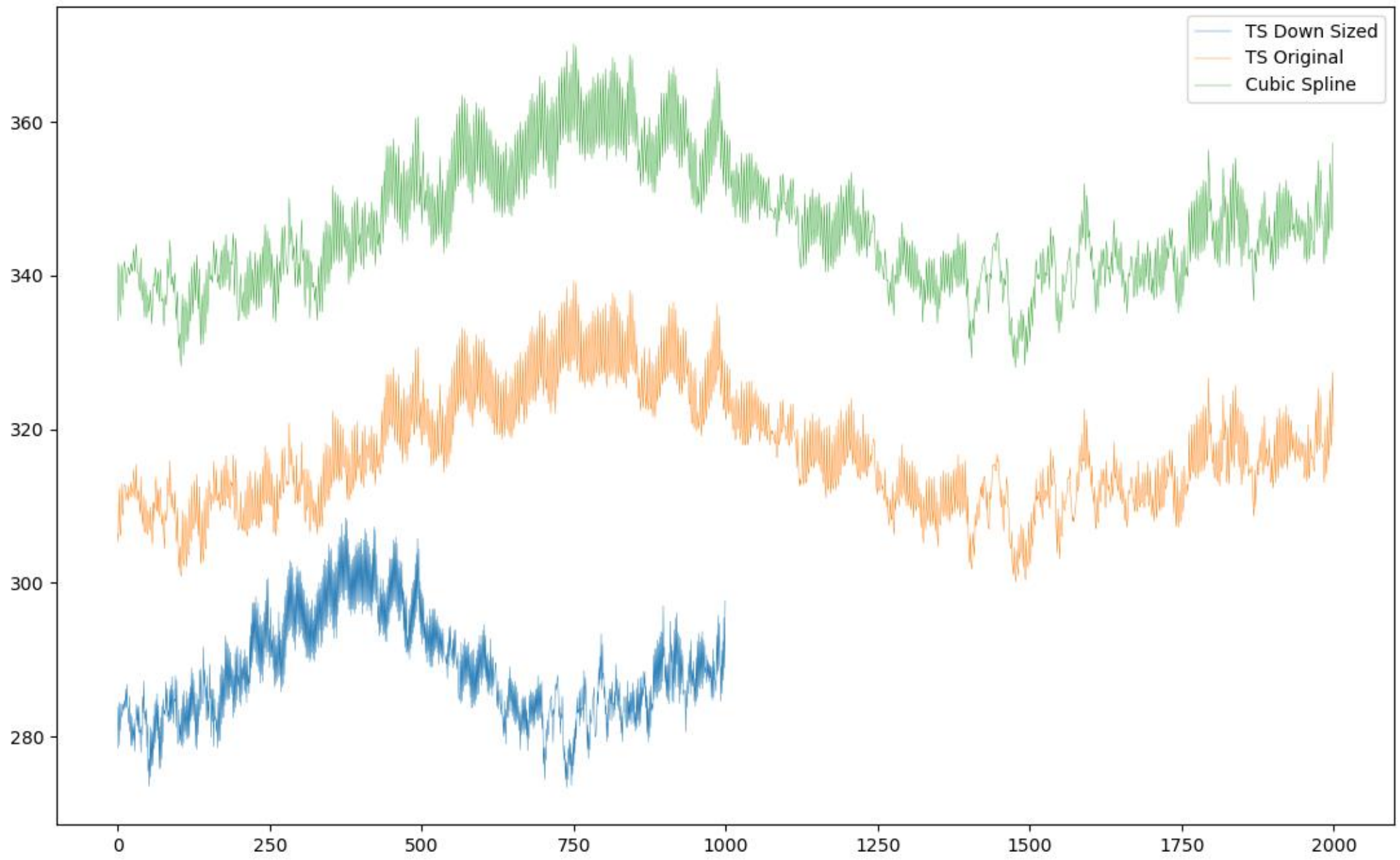

Fig. 3 Air temperature time series reduced uniformly

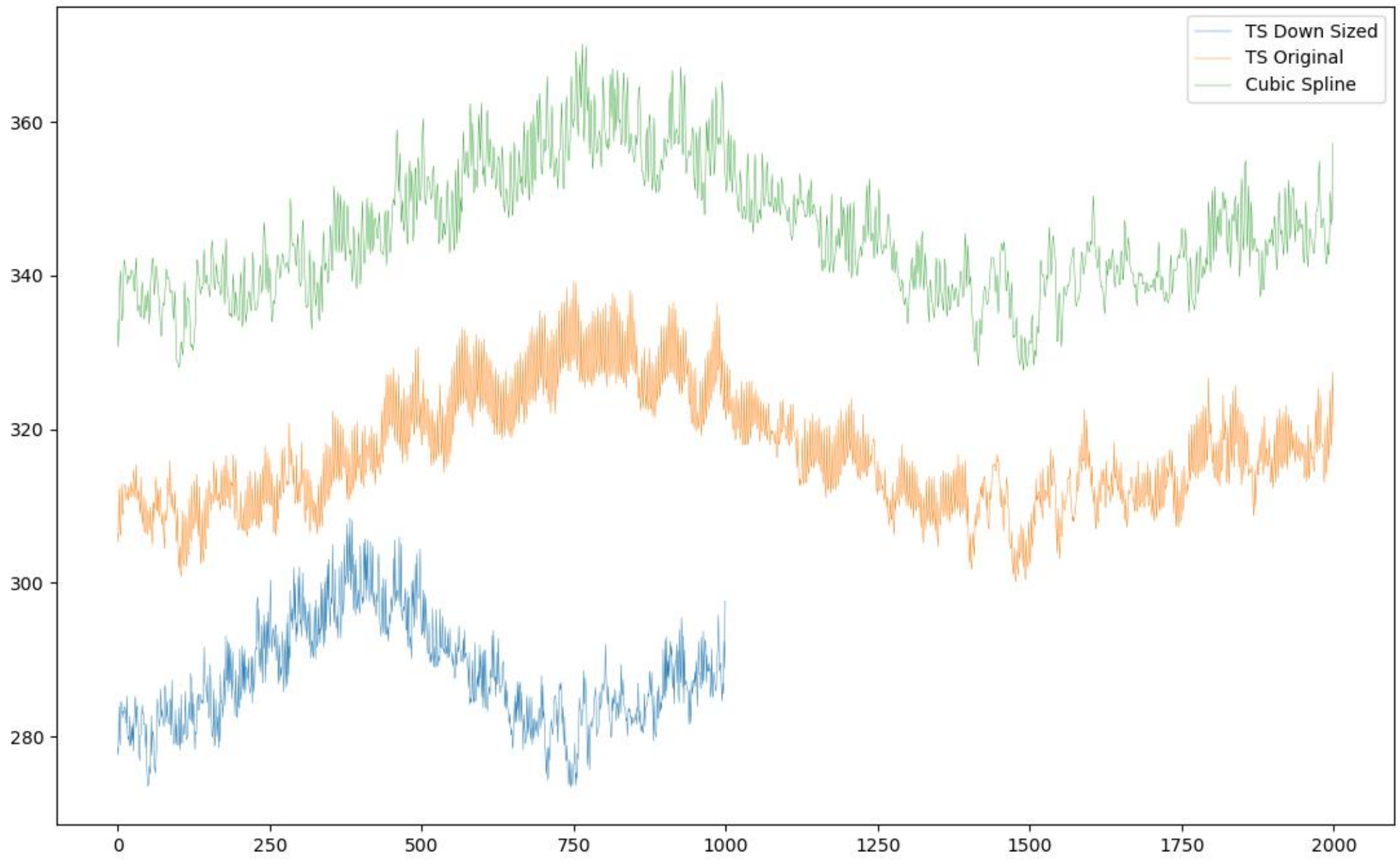

Fig. 4 Air temperature time series reduced randomly

## Discretization

The next step of the methodology is the discretization of the dataset values. This step is fundamental in order to create the LERP-RSA data structure and then execute the ARPaD algorithm in order to detect patterns and cluster the curves. Both LERP-RSA and ARPaD, as already described previously, need to be executed on discrete values. There are several methods for the discretization of continuous values that we can select as already described. Depending on previous experience and/or domain specific attributes, one method may perform better than another. It is up to the domain expert to select the better possibly performed discretization method, although, in practice it has been proven that all methods work more or less equally well in most cases. Still, it is possible to test different discretization methods to check if one performs significantly better than all other. After the standardization of the data values the minimum and maximum values are recorded and the space between splits to classes according to the discretization method. The number of classes will also define the size of the alphabet will be used to reconstruct the discretized time series. For example, if we decide to create 20 classes then the first 20 characters from the Latin alphabet can be used.

It is important to mention here that after the z-scoring and since the standard deviation is one, assuming a Normal distribution, we expect our values to be scattered between -3 and 3, since $6\sigma$ can include 99.9% of the dataset. Therefore, if we split our data to 12 classes using, for example, equal width classes, each class will have size 0.5.

After the determination of the number of classes and alphabet we reconstruct the time series by assigning every continuous, real, value to an element of the alphabet and, thus, creating a discrete sequence. This process gives us the additional advantage of smoothing noisy datasets (Fig. 2), which is more important for shape comparison, despite any possible information loss. The worst case time complexity for the discretization is linear for each sequence or *O(mn)* with regard to the input of *m* sequences of length *n*.

## Mutlivariate LERP-RSA Data Structure

The third step of the process is the creation of the multivariate LERP-RSA data structure. The multivariate variant of LERP-RSA has the same characteristics as the standard LERP-RSA described in the previous section, however, it also holds the index of the sequence (or curve) from which the suffix string has been extracted. Therefore, the LERP-RSA has three columns indicating the suffix string itself, the position that the suffix string exists in the sequence and the identifier of the sequence from which the suffix string has been extracted.

The creation of the multivariate LERP-RSA has two stages. First, we create the LERP-RSA of each sequence. The data structures are merged together and then the new, full, LERP-RSA is lexicographically sorted in ascending order by the suffix string, then the position and finally the sequence. In order to achieve better performance, we can classify the LERP-RSA by the first letter creating as many classes as the size of the alphabet and execute the creation in parallel.

Let's assume that we have three curves, which after their discretization the following sequences have been created:

1) abcdcbabcd
2) abcbdbabcd
3) cbcdbaabcb

| | | | | | | | | |
|---|---|---|---|---|---|---|---|---|
| 1 | 6 | abcd | 2 | 0 | abcbd | 3 | 5 | aabcb |
| 1 | 0 | abcdc | 2 | 6 | abcd | 3 | 6 | abcb |
| 1 | 5 | babcd | 2 | 5 | babcd | 3 | 4 | baabc |
| 1 | 7 | bcd | 2 | 1 | bcbdb | 3 | 7 | bcb |
| 1 | 1 | bcdcb | 2 | 7 | bcd | 3 | 1 | bcdba |
| 1 | 4 | cbabc | 2 | 3 | bdbab | 3 | 8 | cb |
| 1 | 8 | cd | 2 | 2 | cbdba | 3 | 0 | cbcdb |
| 1 | 2 | cdcba | 2 | 8 | cd | 3 | 2 | cdbaa |
| 1 | 3 | dcbab | 2 | 4 | dbabc | 3 | 3 | dbaab |
| | (a) | | | (b) | | | (c) | |

Fig. 5 LERP-RSA for abcdcbabcd (a), abcbdbabcd (b) and cbcdbaabcb (c) for SPL=2 and LERP=5

| | | |
|---|---|---|
| 3 | 5 | a a b c b |
| 3 | 6 | a b c b |
| 2 | 0 | a b c b d |
| 1 | 6 | a b c d |
| 2 | 6 | a b c d |
| 1 | 0 | a b c d c |
| 3 | 4 | b a a b c |
| 1 | 5 | b a b c d |
| 2 | 5 | b a b c d |
| 3 | 7 | b c b |
| 2 | 1 | b c b d b |
| 1 | 7 | b c d |
| 2 | 7 | b c d |
| 3 | 1 | b c d b a |
| 1 | 1 | b c d c b |
| 2 | 3 | b d b a b |
| 3 | 8 | c b |
| 1 | 4 | c b a b c |
| 3 | 0 | c b c d b |
| 2 | 2 | c b d b a |
| 1 | 8 | c d |
| 2 | 8 | c d |
| 3 | 2 | c d b a a |
| 1 | 2 | c d c b a |
| 3 | 3 | d b a a b |
| 2 | 4 | d b a b c |
| 1 | 3 | d c b a b |

Fig. 6 Multivariate LERP-RSA

For the specific example we will use as initial SPL and LERP values two and five correspondingly. First, we create the LERP-RSA for each one of the sequences for the specific SPL and LERP values (Fig. 5). For each one of the individual LERP-RSA the first column is the sequence index, the second is the suffix string position and the third column is the actual suffix string. Each LERP-RSA is lexicographically sorted and then combined to a single, multivariate, LERP-RSA data structure (Fig. 6). Of course, we can perform a single sorting after the combination of the three LERP-RSA data structures. The worst case time complexity for the LERP-RSA construction is log-linear for each sequence or *O(mnlogn)* with regard to the input of *m* sequences of length *n*, while the space complexity is also *O(mnlogn).*

## ARPaD Algorithm

In [24] where the BD2C method first introduced, the ARPaD algorithm is executed with both SPL and LERP values set to one (1). The reason to set LERP to value one is that the BD2C compares values having the same x-axis value. Therefore, it identifies only patterns with length one that occur at the same time or same x-axis position. Furthermore, BD2C requires two additional, temporary data structures, i.e., a set and a dictionary. The set was named TempSet and holds the sequence indexes of the suffix strings found to occur at the same position while the dictionary was named TempDict and it is used to hold the frequency of each set. The two additional data structures were used in meta-analysis to determine similarity measures among curves.

The new Curve Clustering using Common Patterns (3CP) methodology described here uses a more advanced approach. In this case the algorithm is executed using as LERP value the theoretical value that is calculated by the Probabilistic Existence of Longest Expected Repeated Pattern Lemma and, furthermore, it uses as SPL any value smaller than LERP. This allows us to detect every repeated pattern, with length from SPL up to LERP, that exists between all sequences. The outcome of the ARPaD is a list of patterns with the additional information of the sequence and the position in the sequence that exists. When ARPaD finishes we can use the results to identify the similarities of sequences based on the common patterns that exist between them using a meta-analysis process described in the next section. This approach gives us some advantages comparing to the simple BD2C methodology. More precisely, by using a long SPL value we define a minimum length of partial similarity among the curves and, therefore, a stronger commonality or similarity measure between sequences. This means that we can avoid single value similarities that occur with gaps and might be the result of, e.g., noise. SPL guarantees that common patterns are long enough to eliminate such abnormal behavior and avoid categorizing sequences as similar.

Examining the example discussed in the previous section, we execute ARPaD algorithm on the LERP-RSA of Fig. 6. ARPaD algorithm will detect all repeated patterns that exist with length from two up to five (Fig. 7). The worst case time complexity for the ARPaD is log-linear or *O(mnlogn)* with regard to the input of *m* sequences of length *n*.

| | | |
|---|---|---|
| 3 | 5 | a a b c b |
| 3 | 6 | a b c b |
| 2 | 0 | a b c b d |
| 1 | 6 | a b c d |
| 2 | 6 | a b c d |
| 1 | 0 | a b c d c |
| 3 | 4 | b a a b c b |
| 1 | 5 | b a b c d |
| 2 | 5 | b a b c d |
| 3 | 7 | b c b |
| 2 | 1 | b c b d b |
| 1 | 7 | b c d |
| 2 | 7 | b c d |
| 3 | 1 | b c d b a a b c b |
| 1 | 1 | b c d c b |
| 2 | 3 | b d b a b |
| 3 | 8 | c b |
| 1 | 4 | c b a b c |
| 3 | 0 | c b c d b a a b c b |
| 2 | 2 | c b d b a |
| 1 | 8 | c d |
| 2 | 8 | c d |
| 3 | 2 | c d b a a b c b |
| 1 | 2 | c d c b a |
| 3 | 3 | d b a a b c b |
| 2 | 4 | d b a b c |
| 1 | 3 | d c b a b |

Fig. 7 ARPaD results on LERP-RSA

## Metadata Analysis

Having completed the execution of ARPaD algorithm we need to perform a meta-analysis that will identify similarities between sequences according to the patterns discovered. In order to achieve this, first we perform a sorting on the results either based on pattern size (Fig. 8.a) or pattern position (Fig. 8.b). After this we perform a top down scan of the sorted patterns and check if a pattern occurs more than once at the same position. If this is true it means that more than one sequence shares the same pattern at the same position and, therefore, they have a common pattern or measure of similarity at that position. In order to record these similarity measures, first of all, we create a list of arrays of the same length as the sequences, named Sequence Position Identifiers (SPI), and we fill them with zeros. The reason for this bit values (true/false) array representation for each position is to mark at which positions we have discovered common patterns and avoid count more than once positions that have already been discovered and counted as common. For each common pattern we discover between sequences, we change the value at the specific position (bit value) that the pattern occurs from zero to one for the specific length of the pattern. Furthermore, in a separate Sequence Commonality Matrix (SCM) we record the length of the pattern in the cell denoted by the two sequences. When a new pattern is discovered, first we check if the corresponding cells of the sequence array have any non zero values. If not, it means that the pattern overlaps partially with another pattern discovered earlier. We change the zero values to one and we increase the commonality matrix cell by the appropriate zero values changed to one. We continue this process until we have checked all patterns. Of course, there can be cases where there is a full overlap where in such a case, we completely omit the pattern and continue.

| | | |
|---|---|---|
| ab | 1 | 0 |
| | 2 | 0 |
| | 1 | 6 |
| | 2 | 6 |
| | 3 | 6 |
| ba | 3 | 4 |
| | 1 | 5 |
| | 2 | 5 |
| bc | 1 | 1 |
| | 2 | 1 |
| | 3 | 1 |
| | 1 | 7 |
| | 2 | 7 |
| | 3 | 7 |
| cb | 3 | 0 |
| | 2 | 2 |
| | 1 | 4 |
| | 3 | 8 |
| cd | 1 | 2 |
| | 3 | 2 |
| | 1 | 8 |
| | 2 | 8 |
| db | 3 | 3 |
| | 2 | 4 |
| abc | 1 | 0 |
| | 2 | 0 |
| | 1 | 6 |
| | 2 | 6 |
| | 3 | 6 |
| bab | 1 | 5 |
| | 2 | 5 |
| bcd | 1 | 7 |
| | 2 | 7 |
| | 1 | 1 |
| | 3 | 1 |
| abcb | 2 | 0 |
| | 3 | 6 |
| abcd | 1 | 0 |
| | 1 | 6 |
| | 2 | 6 |
| babc | 1 | 5 |
| | 2 | 5 |
| babcd | 1 | 5 |
| | 2 | 5 |

(a)

| | | |
|---|---|---|
| ab | 1 | 0 |
| ab | 2 | 0 |
| abc | 1 | 0 |
| abc | 2 | 0 |
| abcb | 2 | 0 |
| abcd | 1 | 0 |
| cb | 3 | 0 |
| bc | 1 | 1 |
| bc | 2 | 1 |
| bc | 3 | 1 |
| bcd | 1 | 1 |
| bcd | 3 | 1 |
| cb | 2 | 2 |
| cd | 1 | 2 |
| cd | 3 | 2 |
| db | 3 | 3 |
| ba | 3 | 4 |
| cb | 1 | 4 |
| db | 2 | 4 |
| ba | 1 | 5 |
| ba | 2 | 5 |
| bab | 1 | 5 |
| bab | 2 | 5 |
| babc | 1 | 5 |
| babc | 2 | 5 |
| babcd | 1 | 5 |
| babcd | 2 | 5 |
| ab | 1 | 6 |
| ab | 2 | 6 |
| ab | 3 | 6 |
| abc | 1 | 6 |
| abc | 2 | 6 |
| abc | 3 | 6 |
| abcb | 3 | 6 |
| abcd | 1 | 6 |
| abcd | 2 | 6 |
| bc | 1 | 7 |
| bc | 2 | 7 |
| bc | 3 | 7 |
| bcd | 1 | 7 |
| bcd | 2 | 7 |
| cb | 3 | 8 |
| cd | 1 | 8 |
| cd | 2 | 8 |

(b)

Fig. 8 ARPaD results sorted by pattern length and positions

Continuing the previous example, in Fig. 8, we have all repeated patterns detected sorted by length and positions, where the first column represents patterns, the second sequences and the last positions. Then we start with the first pattern, "*ab*". This pattern has two different occurrences based on positions, two at position zero and three at position six. Therefore, both positions are occurrences among different sequences and should be recorded. First, we mark position zero for sequences one and two on our SPI by turning bit values zero to one for the two sequences at positions zero and one, since it has length two (Fig. 9.a). Then on the SCM we record at the corresponding cells (1, 2) and (2, 1) the value of the length of pattern "*ab*" (Fig. 9.a). Additionally, we can also record the number of common patterns found (Fig. 9.a). We continue with the next position at index 6 for all three sequences and we change first the bit values at SPI and then record the number of patterns and cumulative length (Fig. 9.b and 9.b). For sequences one and two we have two patterns with total length (or bit values) 4 while sequence three has only one pattern common with length two with the other sequences. The next pattern is "*ba*" that occurs once at position four, which we ignore, and two at position five for sequences one and two. Now, since we have already marked bit values six and seven as one because of pattern "*ab*" then we will change only bit value at position five to one (Fig. 9.c). Therefore, at the SCM we will increase pattern count by one but instead of the length two of pattern "*ba*" we will increase cumulative length by one and change it to five (Fig. 9.c). We continue the same process until we have covered all important patterns and we end up to have eight common bits between sequences one and two (at least 80% similarity), six common bits between sequences one and three (at least 60% similarity) and five common bits between sequences two and three (at least50% similarity).

It is important to note that the similarity ratios are just a lower bound of similarity and the actual similar data points may be more with higher percentage score. This happens because we have selected to execute ARPaD with SPL value two, which means that we miss single data points that are similar.

(a) ab({1,2} → 0)

| | |
|---|---|
| 1 | 1100000000 |
| 2 | 1100000000 |
| 3 | 0000000000 |

| | 1 | 2 | 3 |
|---|---|---|---|
| 1 | | 1(2) | |
| 2 | 1(2) | | |
| 3 | | | |

(b) ab({1,2,3} → 6)

| | |
|---|---|
| 1 | 1100001100 |
| 2 | 1100001100 |
| 3 | 0000001100 |

| | 1 | 2 | 3 |
|---|---|---|---|
| 1 | | 2(4) | 1(2) |
| 2 | 2(4) | | 1(2) |
| 3 | 1(2) | 1(2) | |

(c) ba({1,2} → 5)

| | |
|---|---|
| 1 | 1100011100 |
| 2 | 1100011100 |
| 3 | 0000001100 |

| | 1 | 2 | 3 |
|---|---|---|---|
| 1 | | 3(5) | 1(2) |
| 2 | 3(5) | | 1(2) |
| 3 | 1(2) | 1(2) | |

(d) bc({2,3} → 1)

| | |
|---|---|
| 1 | 1110011100 |
| 2 | 1110011100 |
| 3 | 0110001100 |

| | 1 | 2 | 3 |
|---|---|---|---|
| 1 | | 4(6) | 2(4) |
| 2 | 4(6) | | 2(4) |
| 3 | 2(4) | 2(4) | |

(e) bc({1,2,3} → 7)

| | |
|---|---|
| 1 | 1110011110 |
| 2 | 1110011110 |
| 3 | 0110001110 |

| | 1 | 2 | 3 |
|---|---|---|---|
| 1 | | 5(7) | 3(5) |
| 2 | 5(7) | | 3(5) |
| 3 | 3(5) | 3(5) | |

(f) cd({1,3} → 2)

| | |
|---|---|
| 1 | 1111011110 |
| 2 | 1110011110 |
| 3 | 0111001110 |

| | 1 | 2 | 3 |
|---|---|---|---|
| 1 | | 5(7) | 4(6) |
| 2 | 5(7) | | 3(5) |
| 3 | 4(6) | 3(5) | |

(g) cd({1,2} → 8)

| | |
|---|---|
| 1 | 1111011111 |
| 2 | 1110011111 |
| 3 | 0111001110 |

| | 1 | 2 | 3 |
|---|---|---|---|
| 1 | | 6(8) | 4(6) |
| 2 | 6(8) | | 3(5) |
| 3 | 4(6) | 3(5) | |

Fig. 9 Sequence Position Identifiers and Sequence Commonality Matrix

What is important to mention is that the aforementioned process can be executed with the patterns sorted by their position, as exist in Fig. 8.b, where we can have a different arrangement and scan the patterns based on the positions that appear, from zero to $n$ (where $n$ the length of sequences), and their length. This approach has the benefit to determine easier overlapping patterns with the same starting position. For example, pattern "*ab*" appears at position zero for sequences one and two, while pattern "*abc*", which overlaps "*ab*", appears also at position zero for sequences one and two. The same happens for "*ab*" and "*abc*" at position 6 for all sequences. Therefore, we can skip checking pattern "*ab*" and record directly pattern "*abc*". This is another flexibility of the knowledge of all repeated (common) patterns.

Both approaches have some additional benefits than the simple BD2C. One is that we can identify regions that sequences have better similarity measure than others. For example, two curves might be 50% similar using any similarity method, yet, they are 90% the same at the first half while only 10% at the second half. Using the aforementioned meta-analysis, we can easily detect this with the use of the SPI and an additional index in the SCM like the common patterns counter. This can also be achieved through the position sorted patterns. If, for example, we need to check only the first half of the sequences then we run the meta-analysis through all patterns with starting positions less than half the length of sequence. Another advantage is that if we just care to find similarities and cluster curves based on a specific threshold, we can stop examining sequences that have already reach that threshold or completely ignore curves that have low scores and it is impossible to meet the threshold by any further examination. Therefore, we can stop the meta-analysis without running it through the whole ARPaD results list. Moreover, this process can be executed in parallel by splitting the ARPaD results to smaller segments based on patterns positions, perform the aforementioned meta-analysis and then combine the results. Another advantage is that we can perform the analysis based on different pattern lengths. More complicated meta-analyses could also be executed such as to examine only patterns that have specific characteristics, for example, specific lengths or special behavior, e.g., stable patterns denoted by a single repeated character regardless length ("*aaa*", "*bbbbb*", "*ababab*" etc.).

# Conclusions

In the current paper the new methodology 3CP has been presented that uses text mining and pattern detection approaches in order to cluster time series and in general curves based on their shape by identifying common patterns between curves. The methodology uses a powerful data structure, the LERP-RSA, specifically designed for text mining and pattern detection and a unique algorithm, the ARPaD, in order to achieve the best possible outcome. Furthermore, with the use of LERP-RSA classification and parallel execution of ARPaD on distributed infrastructure we can achieve optimal performance regardless the number of curves and their size. This methodology can be used both for clustering purposes and for the comparison between different time series and curves. Moreover, although ARPaD detects all patterns with length between SPL and LERP, with the use of different, initial parameters for SPL (smaller) and LERP (larger), more patterns will be revealed, giving the possibility to identify similarities between curves at different time intervals. The new approach presented here can also be applied under extreme data noise presence and provide the best possible clustering. Furthermore, the worst case time complexity of the methodology is log-linear $O(mnlogn)$ with regard to the input of $m$ sequences of length $n$ while the same stands for space complexity $O(mnlogn)$. The complexity of the methodology, with regard to its accuracy, flexibility and additional features, it is proved to be one of the best according to the Aghabozorgi, Shirkhorsidi and Wah literature review on time series clustering [1]. Finally, the proposed methodology can be expanded to work with any kind of discrete sequences such as DNA or RNA sequences in bioinformatics.